\documentclass[12pt,twoside]{report}
 \pdfoutput=1 
\usepackage[numbers]{natbib}
\usepackage{amsmath}
\usepackage[ruled,vlined]{algorithm2e}
\usepackage[noend]{algpseudocode}
\usepackage{float}
\usepackage{listings}
\usepackage{xcolor}
\restylefloat{table}

\DeclareMathOperator*{\argmin}{arg\!\min}

\newcommand{\reportauthor}{Seyedeh Niusha Alavi Foumani}
\newcommand{\supervisor}{Professor Wayne Luk}
\newcommand{\degreetype}{Advanced Computing}

\usepackage[a4paper,hmargin=2.8cm,vmargin=2.0cm,includeheadfoot]{geometry}
\usepackage{textpos}
\usepackage{textcomp}
\usepackage{tabularx,longtable,multirow,subfigure,caption}
\usepackage{fancyhdr} 
\usepackage{url} 
\usepackage[english]{babel}
\usepackage{amsmath}
\usepackage{graphicx}
\usepackage{dsfont}
\usepackage{epstopdf} 
\usepackage{backref} 
\usepackage{array}
\usepackage{latexsym}
\usepackage[pdftex,pagebackref,hypertexnames=false,colorlinks]{hyperref} 

\hypersetup{pdftitle={},
  pdfsubject={}, 
  pdfauthor={},
  pdfkeywords={}, 
  pdfstartview=FitH,
  pdfpagemode={UseOutlines},
  bookmarksnumbered=true, bookmarksopen=true, colorlinks,
    citecolor=black,%
    filecolor=black,%
    linkcolor=black,%
    urlcolor=black}

\usepackage[all]{hypcap}

\def\@makechapterhead#1{%
  \vspace*{10\p@}%
  {\parindent \z@ \raggedright \sffamily
    \interlinepenalty\@M
    \Huge\bfseries \thechapter \space\space #1\par\nobreak
    \vskip 30\p@
  }}

\def\@makeschapterhead#1{%
  \vspace*{10\p@}%
  {\parindent \z@ \raggedright
    \sffamily
    \interlinepenalty\@M
    \Huge \bfseries  #1\par\nobreak
    \vskip 30\p@
  }}

\allowdisplaybreaks

\date{April 2020}

\begin{document}


\begin{titlepage}
\newcommand{\HRule}{\rule{\linewidth}{0.5mm}} 


\includegraphics[width = 4cm]{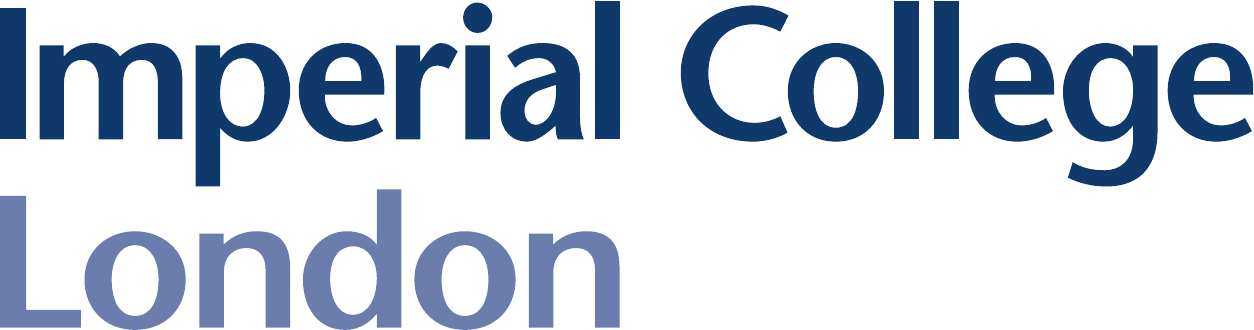}\\[0.5cm] 

\center 


\textsc{\Large Imperial College London}\\[0.5cm] 
\textsc{\large Department of Computing}\\[0.5cm] 


\HRule \\[0.4cm]
{ \Large \bfseries  An Analysis of Alternating Direction Method of Multipliers for Feed-forward Neural Networks\\[0.3cm] } 
\HRule \\[1.5cm]
 

\begin{minipage}{0.4\textwidth}
\begin{flushleft} \large
\emph{Author:}\\
\reportauthor 
\end{flushleft}
\end{minipage}
~
\begin{minipage}{0.4\textwidth}
\begin{flushright} \large
\emph{Supervisor:} \\
\supervisor 
\end{flushright}
\end{minipage}\\[4cm]

\vfill 
Submitted in partial fulfillment of the requirements for the MSc degree in
\degreetype~of Imperial College London\\[0.5cm]

\makeatletter
\@date 
\makeatother

\end{titlepage}

\pagenumbering{roman}
\clearpage{\pagestyle{empty}\cleardoublepage}
\setcounter{page}{1}
\pagestyle{fancy}

\begin{abstract}
In this work, we present a hardware compatible neural network training algorithm in which we used alternating direction method of multipliers (ADMM) and iterative least-square methods. The motive behind this approach was to conduct a method of training neural networks that is scalable and can be parallelised. These characteristics make this algorithm suitable for hardware implementation. 
We have achieved 6.9\% and 6.8\% better accuracy comparing to SGD and Adam respectively, with a four-layer neural network with hidden size of 28 on HIGGS dataset. Likewise, we could observe 21.0\% and 2.2\% accuracy improvement comparing to SGD and Adam respectively, on IRIS dataset with a three-layer neural network with hidden size of 8. This is while the use of matrix inversion, which is challenging for hardware implementation, is avoided in this method. We assessed the impact of avoiding matrix inversion on ADMM accuracy and we observed that we can safely replace matrix inversion with iterative least-square methods and maintain the desired performance. Also, the computational complexity of the implemented method is polynomial regarding dimensions of the input dataset and hidden size of the network.

\end{abstract}

\cleardoublepage
\section*{Acknowledgements}
I would like to extend my sincere gratitude to my supervisor, Prof. Wayne Luk and Dr. Ce Guo for their valuable help and advice.

\clearpage{\pagestyle{empty}\cleardoublepage}

\fancyhead[RE,LO]{\sffamily {Table of Contents}}
\tableofcontents

\clearpage{\pagestyle{empty}\cleardoublepage}
\pagenumbering{arabic}
\setcounter{page}{1}
\newcommand{\changefont}{%
    \fontsize{11}{11}\selectfont}
\fancyhead[LE,RO]{\changefont\slshape \rightmark \tiny}
\fancyhead[LO,RE]{\changefont\slshape \leftmark \small}

\chapter{Introduction}
Neural networks are increasingly being used for solving problems in various fields. This is while the amount of available data to train our models is exploding, and the architecture of neural networks are becoming more and more complex. As a result, the use of hardware-accelerated and scalable methods in this field is drawing more attention than ever \cite{schuman2017survey}. \vspace{8pt}\\
The main obstacle for the use of hardware accelerators like GPUs and FPGAs is that most of the currently used methods for training neural networks are not ideal for hardware implementation. This is mainly due to the following facts:
\begin{itemize}
    \item These algorithms are sequential in principle and suffer from a strong sequential dependency.
    
    \item They may include operations which are expensive for hardware. One example of such operations could be matrix inversion
\cite{kumar2014approach}.
\end{itemize}
In this project, we implemented an algorithm to train neural networks using alternating direction method of multipliers (ADMM).  In order to make this method more hardware-friendly, we used an iterative least-square method to avoid computing matrix inversion. The contributions of our work can be described as below:
\begin{itemize}
    \item Complexity analysis of ADMM and theoretical proof of one of the main procedures.
    \item The use of LSMR \cite{fong2011lsmr} as an iterative least-square solver, which can be tuned faster than closed-form solvers and is a big step toward hardware implementation.
    \item Comparison between the implemented ADMM-based method and gradient-based methods (SGD and Adam).
\end{itemize}
 \vspace{8pt}
In chapter \ref{ch2} an overview of artificial neural networks, gradient-based methods and mathematical concepts of ADMM are described. In section \ref{ADMMNNsection} the method of training neural networks using ADMM proposed in \cite{taylor2016training} has been discussed in more details. Later in chapter \ref{ch3} we elaborate the use of iterative least-square methods in our implementation alongside theoretical and complexity analysis. 
In chapter \ref{ch4} an experimental comparison of the implemented method with two-gradient based methods is presented. Finally, chapter \ref{ch5} includes a conclusion of this report and the possibilities for future works.


\chapter{Background}
\label{ch2}
\section{Artificial Neural Networks}
\label{ANN}
\begin{figure}[tb]
\centering
\includegraphics[width = 0.6\hsize]{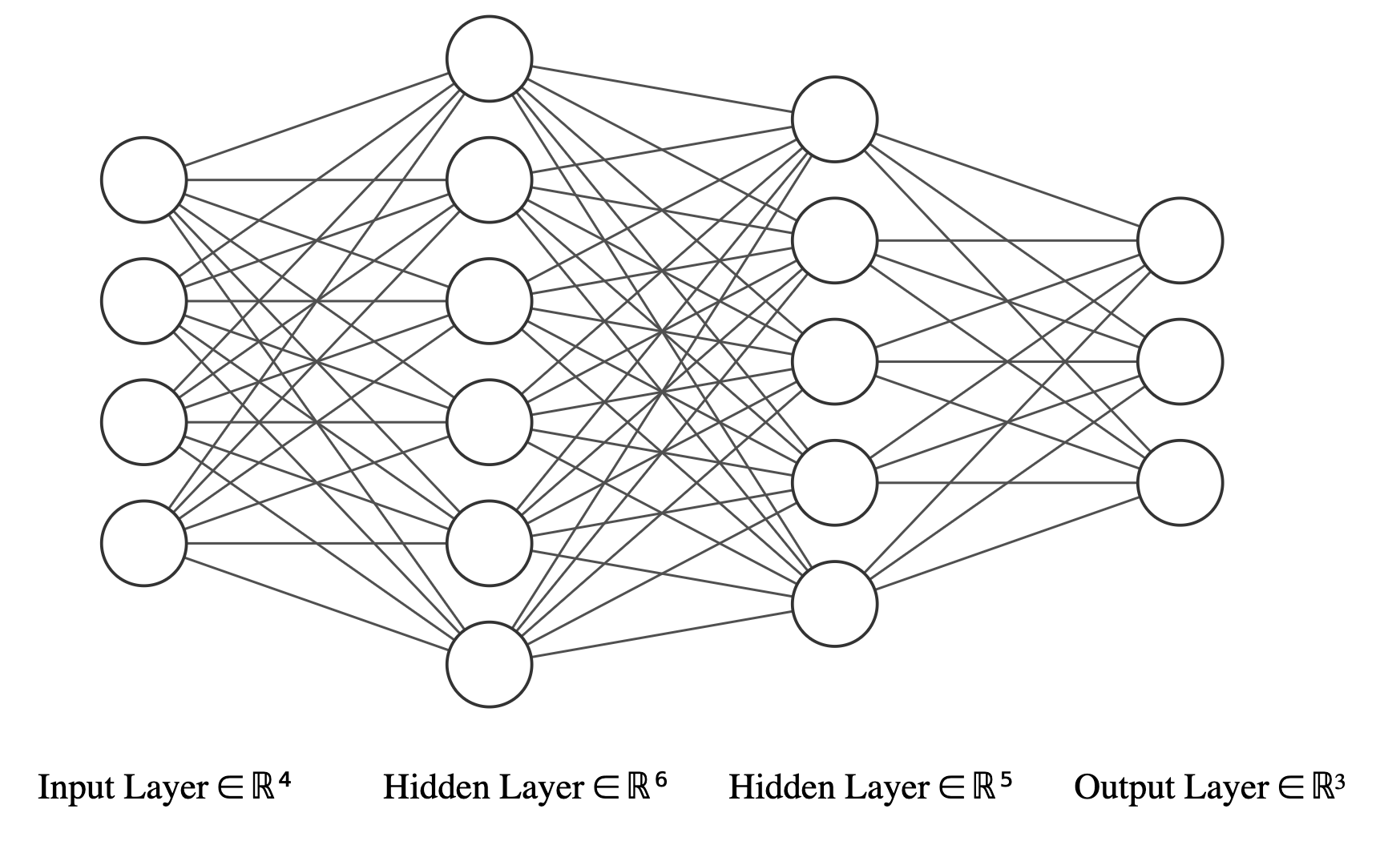}
\caption{A simple feed forward neural network }
\label{fig:nn}
\end{figure}

The primary purpose of most of ML algorithms is to find the best approximation of some function $f^*$ by learning the optimal parameters. In the artificial neural networks, these learnable parameters are the weight matrices which along with activation functions form the neurons. Activation functions usually are used to introduce non-linearity to neural networks, but they can also be linear functions \cite{Deep}. In feed-forward neural networks, a group of neurons that have the same input but different weights construct a layer and a collection of layers are chained together as shown in the figure \ref{fig:nn}. Following statements hold for a simple three-layer neural network with $N$ training samples, $D$ features, $HS$ number of neurons in the hidden layer and $OS$ as the dimensions of output. $h_l$ is the activation function of layer l. (This notation is used for neural networks in following sections of this report)
\begin{gather*}
    \text{Input data }\hspace{5pt}  x_{0} \in {\rm I\!R}^{D*N}  \\
    W_1 \in {\rm I\!R}^{HS*D} \\  z_1= W_1x_0 ,\hspace{5pt} z_1\in {\rm I\!R}^{HS*N}  \\
     \text{Input of hidden layer }\hspace{5pt} x_1 = h_1(z_1) \in {\rm I\!R}^{HS*N}  \\
     W_2 \in {\rm I\!R}^{HS*HS}
      \\
     z_2= W_2x_1 ,\hspace{5pt} z_2\in {\rm I\!R}^{HS*N}  \\
    \text{Input of last layer}\hspace{5pt} x_2 = h_2(z_2) \in {\rm I\!R}^{HS*N} \\ W_3 \in {\rm I\!R}^{OS*HS}
    \\
    \text{Output}\hspace{5pt} z_3= W_3x_2 ,\hspace{5pt} z_3\in {\rm I\!R}^{OS*N} 
\end{gather*} 
By defining a loss function $\ell$, we can consider training a neural network as an optimisation problem: 
\begin{gather}
    \label{NNopt}
    \min_{W} \ell(f(x_0,W), y) 
\end{gather}
where $W$ (weight matrices) is the parameter that we want to learn such that the output of the function $f$ given the input $x_0$ be as close as possible to the actual output $y$.

\begin{figure}[tb]
\centering
\includegraphics[width = 0.5\hsize]{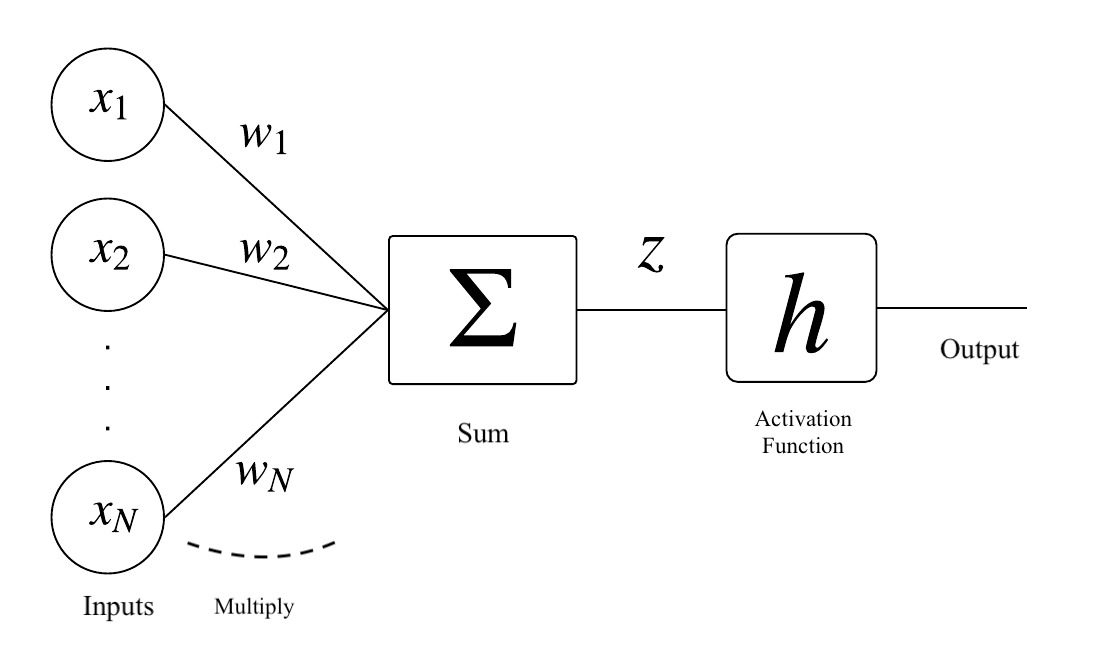}
\caption{Internal structure of a single neuron}
\label{fig:neuron}
\end{figure}

\section{Gradient-Based Methods and Back-propagation }
 Gradient-based methods are iterative optimisation algorithms that use the first-order information derived from the objective function. In each iteration of these algorithms, a feasible direction and step size are chosen to move towards a more optimal solution.\vspace{8pt}\\
Gradient-based methods \cite{ruder2016overview} along with back-propagation \cite{rumelhart1986learning} as a gradient computing technique are widely used to train the neural networks. These methods suffer from several fundamental problems which are explored in the following paragraphs.\vspace{8pt}\\
The problem of vanishing and exploding gradients occur as a result of repeated matrix multiplications. 
Multiplying small values of gradient multiple times makes the gradient very small. This could slow down or in some cases stop the process of learning. 
On the other hand, multiplying big values of gradients multiple times would lead to exploding gradients and make the learning process unstable.  This problem is crucial especially in RNN networks \cite{bengio1994learning}. Several approaches have been proposed to reduce the effect of this problem. For example, using rectified linear units (ReLU) \cite{nair2010rectified} as activation function and changing the architecture to Long Short Term Memory (LSTM) networks \cite{hochreiter1997long} can help to mitigate vanishing gradients. Clipping gradient could also be a solution for exploding gradients.\vspace{8pt}\\
The other problem is the lack of scalability which stems from the nature of gradient-based algorithms and their sequential dependency. In gradient-based methods, the weights are updated after the gradient of a mini-batch becomes available. Gradient computation for a new mini-batch cannot start before the previous one finishes.\vspace{8pt}\\
Since the optimisation problem of neural networks is usually non-convex, converging to local minima or saddle points is another issue of gradient based methods. Also, it has been shown that saddle points become a more critical issue in the higher dimensions \cite{dauphin2014identifying}. \vspace{8pt}\\
Gradient based method are also sensitive to ill-conditioning \cite{wang2019admm} \cite{chong1996chong}.\vspace{8pt}\\
Up to this point, many state-of-the-art variants of the gradient based methods have been devised including SGD \cite{bottou2012stochastic}, AdaGrad \cite{duchi2011adaptive}, AdaDelta \cite{zeiler2012adadelta} and the most popular Adam \cite{kingma2014adam}. But in general it is observed that these methods which use back propagation technique are usually inconsistent and have unpredictable performance \cite{white1987some}. 
\section{Neural Networks In Hardware}

As we mentioned before the growing interest in neural networks and the rapidly growing amount of available data has urged the use of hardware acceleration in this field. There are several approaches to alter the process of training neural networks and make them more hardware compatible. Currently, the main solutions are around the following three ideas \cite{sze2017hardware}:
\begin{itemize}
    \item Reduce Precision: It is possible to use fixed-point or low precision numbers in both training and inference part of neural networks. Works in this area are mostly concentrated around the inference. Using this approach in training neural networks with gradient-based methods usually worsens the accuracy \cite{fox2019training} \cite{courbariaux2016binarynet} \cite{wang2018training}.
    \item Sparsity: This technique suggests to prune some weights of the neural network. This can be achieved either by eliminating the weights that have less impact on output or have a small value \cite{sze2017hardware}. 
    \item Compression: This idea is about compressing the data in order to reduce data movement and storage cost \cite{sze2017hardware}.
\end{itemize}
\newpage
\section{Alternating Direction
Method of Multipliers }
Alternating Direction Method of Multipliers (ADMM) \cite{gabay1976dual}  is a powerful optimisation method that breaks convex problems into smaller subproblems in order to solve them \cite{boyd2011distributed}. The advantage of this method is that it can be used in large-scale problems in the machine learning and statistic fields and it also has the potential of being applied in parallel.\vspace{8pt}\\
Dual decomposition and method of multipliers are two important algorithms that are considered to be ancestors of the ADMM. In this section first a simple overview of these two optimisation algorithms is given, then the ADMM algorithm itself is elaborated.
\subsection{Dual Decomposition}
Consider the following optimisation problem: 
\begin{gather}
\label{opt}
    \min f(x)
    \\ 
    \text{subject to   } Ax=b \nonumber
\end{gather}
Where $x \in {\rm I\!R}^{n}$, $A \in {\rm I\!R}^{m*n}$, $b \in {\rm I\!R}^{m}$ and $f$ from ${\rm I\!R}^{n}$ to $ {\rm I\!R}$  is a convex function.\\
The Lagrangian function associated with the optimisation problem \ref{opt} is: 
\begin{gather}
    L(x,\lambda) = f(x) + \lambda^T(Ax-b)
\end{gather}
Where $\lambda \in {\rm I\!R}^{m}$. In order to perform dual ascent method on \ref{opt} we have to use gradient ascent to solve the dual problem \ref{dual}.
\begin{gather} \label{dual}
\max g(\lambda)
\\ 
g(\lambda) = inf_x L(x,\lambda)
\end{gather}
Where $g(\lambda)$ is the dual function.\vspace{8pt}\\
Assuming that $L(x,\lambda^*)$ has only one minimiser ($f$ is strictly convex), the primal optimal point $x^*$ is calculated using the dual optimal point $\lambda^*$: 
\begin{gather}
x^* = \argmin_x L(x,\lambda^*)
\end{gather}
To apply gradient ascent on the dual problem we have to iteratively update $\lambda$. Assuming that $g$ is differentiable we can write: 
\begin{gather}
    \lambda^{k+1} = \lambda^k + \alpha^k \nabla g(\lambda^k)
\end{gather}
Where $\alpha^k > 0 $ is the step size at iteration $k$. 
\newpage
We can compute $\nabla g(\lambda)$ from the following: 

\begin{gather}
    \nabla g(\lambda^k) = A\hat{x} - b
\\
     \hat{x}  = \argmin_x L(x,\lambda^k)
\end{gather}
In summary, the dual ascent method is two iterating updates: 
\begin{gather}
    \label{xmin}
    x^{k+1} = \argmin_x L(x,\lambda^k)
    \\
    \lambda^{k+1} = \lambda^k + \alpha^k(Ax^{k+1}-b)
\end{gather}
Now assume that the objective function $f$ is separable: 
\begin{gather}
    f(x) = \sum_{i=1}^N f_i(x_i)
\\
    x = (x_1,...,x_N)
\end{gather}
Then the Lagrangian function is also separable in $x$: 
\begin{gather}
    L(x,\lambda) =  \sum_{i=1}^N L_i(x_i,\lambda)  \\
     L_i(x_i,\lambda) = f_i(x_i) + \lambda^TA_ix_i - (1/N)\lambda^Tb
\\
    A = [A_1,...,A_N]
\end{gather}
In this case, we can split \ref{xmin} into $N$ minimisations that are independent and can be computed in parallel. This algorithm is called dual decomposition \cite{everett1963generalized}: 
\begin{gather}
    x_i^{k+1} = \argmin_{x_i} L_i(x_i,\lambda^k), \hspace{5pt} i = 1,...N
    \\
    \lambda^{k+1} = \lambda^k + \alpha^k(Ax^{k+1}-b)
\end{gather}
This algorithm can be used to solve large problems with numerous strong assumptions. 

%
%
\subsection{Method of Multipliers}
Applying the dual ascent method on the augmented Lagrangian of an optimisation problem is called method of multipliers \cite{hestenes1969multiplier} . Augmented Lagrangian methods are used to make the dual ascent algorithm converge under milder assumptions (to be specific we can eliminate the assumption of convexity of $f$ ).
\vspace{5pt}\\
The augmented Lagrangian associated with the optimisation problem \ref{opt} is: 
\begin{gather}
    L_p(x,\lambda) = f(x) + \lambda^T(Ax-b) + (p/2)||Ax-b||_2^2
\end{gather}
Where $p>0$ is the penalty term.
The augmented Lagrangian of \ref{opt} can be seen as the standard Lagrangian for the equivalent problem \ref{equi}:
\begin{gather}
\label{equi}
    \min f(x) +(p/2)||Ax-b||_2^2
    \\ 
    \text{subject to   } Ax=b \nonumber
\end{gather}
We can write the dual function:
\begin{gather}
    \label{pdual}
    g_p(\lambda) = inf_x L_p(x,\lambda)
\end{gather}
\ref{pdual} can be shown to be differentiable under milder assumptions compare to the original problem. By applying the dual ascent method with step size equal to $p$ we have: 
\begin{gather}
    \label{xminp}
    x^{k+1} = \argmin_x L_p(x,\lambda^k)
    \\
    \lambda^{k+1} = \lambda^k + p(Ax^{k+1}-b)
\end{gather}

It worth mentioning that when $f$ is separable we can not conclude that augmented Lagrangian is also separable. Therefore we can not break the minimisation step of the algorithm \ref{xminp} into subproblems that can be solved in parallel.
\subsection{ADMM Algorithm}
Consider the following optimisation problem: 
\begin{gather}
\label{admmopt}
    \min f(x) + g(z)
    \\ 
    \text{subject to   } Ax + Bz = c \nonumber
\end{gather}
Where $x \in {\rm I\!R}^{n}$, $z \in {\rm I\!R}^{m}$, $A \in {\rm I\!R}^{p*n}$, $B \in {\rm I\!R}^{p*m}$, $c \in {\rm I\!R}^{p}$ and $f$ and $g$ are convex functions. The augmented Lagrangian associated with \ref{admmopt} is: 
\begin{gather}
    L_p(x,z,\lambda) = f(x) + g(z)+ \lambda^T(Ax + Bz - c) + (p/2)||Ax + Bz - c||_2^2
\end{gather}
The method of multipliers for \ref{admmopt} can be written as:
\begin{gather}
    \label{jointly}
    (x^{k+1} ,  z^{k+1} ) = \argmin_{x,z} L_p(x,z,\lambda^k)
    \\
     \lambda^{k+1} = \lambda^k + p(Ax^{k+1} + Bz^{k+1} - c )
\end{gather}
In \ref{jointly} we minimise over $x$ and $z$ jointly. In case of ADMM algorithm the minimisation over $x$ and $z$ is separated (minimise over $x$ while holding $z$ fixed and vice versa).
Each iteration of the ADMM algorithm to solve the problem \ref{admmopt} includes three updates:
\begin{gather}
    \label{admm}
    x^{k+1} = \argmin_x L_p(x,z^k,\lambda^k)
    \\
    z^{k+1} = \argmin_z L_p(x^k+1,z,\lambda^k)
    \\
    \lambda^{k+1} = \lambda^k + p(Ax^{k+1} + Bz^{k+1} - c )
\end{gather}
\section{ADMM for Neural Networks }
\label{ADMMNNsection}
ADMM can be used as an optimisation algorithm in neural networks. Our implemented method is based on the presented methodology in \textit{"Training Neural Networks Without Gradients: A Scalable ADMM Approach"}\cite{taylor2016training}. In this section, we briefly describe their work using the notation mentioned in section \ref{ANN}.\vspace{8pt}\\
 The main idea presented in \cite{taylor2016training} is to store the output of each layer $l$, in a variable called pre-activation $z_l$ to be able to apply ADMM to the optimisation problem of a neural network. By applying this technique, we can decouple the weights of the neural network from the activation functions and change the optimisation problem \ref{NNopt}  of an $L$ layer neural network to the following equivalent problem:
\begin{gather}
    \label{admmNN}
    \min_{W_l , x_l,z_l} \ell(z_L, y) 
    \\ 
    \text{subject to  }\hspace{5pt} z_l = W_lx_{l-1}, \text{ for } l = 1,2,...L \nonumber
    \\   x_l = h_l(z_l), \text{ for }l = 1,2,...L-1\nonumber
\end{gather}
The augmented Lagrangian of \ref{admmNN} is:
\begin{gather}
  \ell(z_L, y)  + \beta_L||z_L - W_Lx_{L-1}||_2^2  \\ \nonumber  + \sum_{l=1}^{L-1}[\gamma_l||x_l - h_l(z_l)||_2^2 + \beta_l||z_l - W_lx_{l-1}||_2^2] \\ \nonumber + 
  \sum_{l=1}^{L-1} \lambda_l^T(z_l - W_lx_{l-1}) + \delta_l^T(x_l - h_l(z_l)) \\ \nonumber + 
  \lambda_L^T(z_L - W_Lx_{L-1}) + \delta_L^T(x_L - h_L(z_L)) 
\end{gather}
Where $\gamma_l$ and $\beta_l$ are penalty parameters and $\lambda_l$ and $\delta_l$ are vectors of Lagrangian multipliers.
In \cite{taylor2016training} it is mentioned that applying the classic ADMM to \ref{admmNN} and using a separate Lagrangian vector for each of the constraints would make the algorithm extremely unstable. Their proposed method takes into account just one of the Lagrangian multiplier vectors which yield:
\begin{gather}
    \label{ADMML}
  \ell(z_L, y)  + \beta_L||z_L - W_Lx_{L-1}||_2^2  \\ \nonumber  + \sum_{l=1}^{L-1}[\gamma_l||x_l - h_l(z_l)||_2^2 + \beta_l||z_l - W_lx_{l-1}||_2^2] \\ \nonumber + 
  \lambda^T(z_L - W_Lx_{L-1})
\end{gather}
The only Lagrangian multiplier vector used in \ref{ADMML} is $\lambda$ which has the same dimensions as $z_L$. Pseudo-code of the algorithm can be seen in \ref{alg:algo}. The algorithm moves forward by updating one variable at a time while keeping the others fixed. In the following sections minimisation steps of the algorithm are discussed.

\begin{algorithm}
 \While{not converged}{
  \For { $l =1,2,... L-1$} {
  $W_l \leftarrow z_lx_{l-1}^\dagger$ \\ 
  $x_l  \leftarrow ( \gamma_l +\beta_{l+1}W_{l+1}^T W_{l+1}) ^{-1}(\gamma_lh_l(z_l) + \beta_{l+1}W_{l+1}^T z_{l+1}) $
  \\
  $z_l \leftarrow \argmin_z {\gamma_l||x_l - h_l(z_l)||_2^2 + \beta_l||z_l - W_lx_{l-1}||_2^2 }$

  }
 $ W_L \leftarrow z_Lx_{L-1}^\dagger $\\ 
  $z_L \leftarrow \argmin_z { \ell(z_L, y)  + \beta_L||z_L - W_Lx_{L-1}||_2^2 + \lambda^T(z_L - W_Lx_{L-1})}$\\
  $\lambda \leftarrow \lambda + \beta_L(z_L - W_Lx_{L-1})$

 }
 \caption{ADMM for Neural Networks}
\label{alg:algo}
\end{algorithm}
\newpage
\subsection{Weight Update}
Minimising \ref{ADMML} with respect to $W_l$ is a simple least-square problem with solution :
\begin{gather}
\label{weightupdate}
W_l \leftarrow z_lx_{l-1}^\dagger
\end{gather}
where  $x_{l-1}^\dagger$ is the pseudo-inverse of the matrix $x_{l-1}$.
\subsection{Activation Update}
The new value of $x_l$ in \cite{taylor2016training} is given by:
\begin{gather}
    \label{activationupdate}
    x_l  \leftarrow ( \gamma_l +\beta_{l+1}W_{l+1}^T W_{l+1}) ^{-1}(\gamma_lh_l(z_l) + \beta_{l+1}W_{l+1}^T z_{l+1}) 
\end{gather}
This equation comes without proof in \cite{taylor2016training}. Details of deriving this equation are discussed in section \ref{proof}.
\subsection{Output Update}
In order to find the new value of $z_L$ we have to solve \ref{outputupdate} which is a non-convex and non-quadratic problem because of the activation function $h$. The activation function works element-wise on its inputs, therefore when $h$ is piece-wise linear we can easily solve \ref{outputupdate} in closed form.
\begin{gather}
\label{outputupdate}
    \argmin_z {\gamma_l||x_l - h_l(z_l)||_2^2 + \beta_l||z_l - W_lx_{l-1}||_2^2 }
\end{gather}
\subsection{Lagrangian Multiplier Update}
The equation for updating the value of the Lagrangian multiplier in \cite{taylor2016training} is given by:
\begin{gather}
      \lambda \leftarrow \lambda + \beta_L(z_L - W_Lx_{L-1})
\end{gather}


\chapter{Contribution}
\label{ch3}
In this project, first we implemented the method explained in \ref{ADMMNNsection} as it has a great potential of being implemented in parallel on hardware platforms. Then in order to make the algorithm more feasible to be parallelised on FPGAs, a least-square iterative method was implemented to be used instead of closed-form solvers for matrix inversion. \vspace{8pt}\\
The key characteristics of our implementation can be described as the following:
\begin{itemize}
    \item This method is parallel by nature and does not suffer from the sequential dependency associated with gradient-based methods.
    \item As oppose to gradient-based methods, our method can be combined with using low-precision or fixed-point number technique without its precision being drastically affected. This is due to avoiding back-propagation.
    \item The only obstacle for implementing the ADMM-based training method on hardware is computation of matrix inversion which is avoided in our implemented method.
\end{itemize}

In the following sections, the proof of equation \ref{activationupdate} and complexity analysis of the final implemented method is also provided.

\section{Activation Update Equation Proof}
\label{proof}
We want to find the value of $x_l$ that minimises the objective function, keeping all other variables fixed. The matrix $x_l$ appears in two terms of \ref{ADMML}. The minimisation task is given by:
\begin{gather}
\argmin_{x_l} \beta_{l+1}||z_{l+1} - W_{l+1}x_l||^2 + \gamma_l||x_l -h_l(z_l)||^2  
    \label{minal}
\end{gather}
Where we can show that $g(x_l) = \beta_{l+1}||z_{l+1} - W_{l+1}x_l||^2 + \gamma_l||x_l -h_l(z_l)||^2  $ is a strictly convex function. 
\begin{gather}
\nabla g(x_l) = 2\gamma_l(x_l - h_l(z_l)) - 2\beta_{l+1}W_{l+1}^T(z_{l+1}- W_{l+1}x_l)
    \label{d_al}
\end{gather}
\begin{gather}
\nabla^2 g(x_l) = 2\gamma_lI + 2\beta_{l+1}W_{l+1}^T  W_{l+1}     \label{d2_al}
\end{gather}
It can be seen that $\nabla^2 g(x_l)$ is positive definite, so $g(x_l)$ is strictly convex and we can find the global minimiser by forcing the first-order derivative to be equal to zero.
\begin{gather}
 \gamma_l(x_l - h_l(z_l)) - \beta_{l+1}W_{l+1}^T(z_{l+1}- W_{l+1}x_l) = 0
 \nonumber \\
  \gamma_la_l - \gamma_lh_l(z_l) - \beta_{l+1}W_{l+1}^T z_{l+1}+\beta_{l+1}W_{l+1}^T W_{l+1}x_l = 0
   \nonumber \\
  ( \gamma_l +\beta_{l+1}W_{l+1}^T W_{l+1}) x_l =  \gamma_lh_l(z_l) + \beta_{l+1}W_{l+1}^T z_{l+1}
  \\
  x_l = ( \gamma_l +\beta_{l+1}W_{l+1}^T W_{l+1}) ^{-1}(\gamma_lh_l(z_l) + \beta_{l+1}W_{l+1}^T z_{l+1}) \nonumber
\end{gather}

\section {Using a Least-Square Iterative Method}
In \cite{taylor2016training} for training a neural network with one hidden layer and hidden size equal to 300, 7200 cores have been used. The most computationally expensive part of the ADMM algorithm originates from the matrix inversion in parameter updates \cite{wang2019admm}. The time complexity of computing a matrix inversion is approximately $O(n^3)$, where $n$ is the dimension of the rectangular matrix.\vspace{8pt}\\
To avoid performing matrix inversion we implemented a fast iterative least-square solver, LSMR \cite{fong2011lsmr}.
This method has the potential of becoming faster by relaxing the convergence conditions. One approach could be reducing the number of the main loop iterations and limit it to a constant in order to reduce time complexity.
The other advantage of this method is that it can be parallelised and be implemented on hardware platforms which is our final goal.\vspace{8pt}\\
In our implemented method, an iterative least-square solver is utilised in computing the new values of $w_l$ and $x_l$ in equations 
\ref{weightupdate} and \ref{activationupdate} respectively.
\section{Time Complexity Analysis }
\label{complexity}
In this section we provide time complexity of each procedure in our implementation.
\begin{itemize}

\item \textbf{Complexity of LSMR Implementation}:\vspace{5pt}\\
The implemented LSMR function takes a matrix $A \in {\rm I\!R}^{m*n}$ and a vector $b \in {\rm I\!R}^{m}$ as input. The computational complexity of this function comes from the Golub-Kahan bidiagonalization process \cite{golub1965calculating}. The implementation contains a main loop that iterates $min(m,n)$ times. The complexity of each iteration comes from the dot product between the matrix $A$ and a vector of size $n$. Therefore the time complexity of this function is $O( min(m,n) * m * n )$ .

\item \textbf{Complexity of \textit{Weight Update} Procedure:}\vspace{5pt}\\
The weight update function takes two matrices $z_l \in {\rm I\!R}^{m*n}$ and  $x_{l-1} \in {\rm I\!R}^{p*n}$. We call the LSMR function $m$ times, each time with $x_{l-1}^T$ and a column of $z_{l}$ as inputs. Hence the time complexity of this function would be $O( min(n,p) * n * p * m)$ . When we call this function in hidden layers we have $ m = p = HS$.

\item \textbf{Complexity of \textit{Activation Update} Procedure:} \vspace{5pt}\\
This function takes five inputs. Three of them are matrices: $W_{l+1} \in {\rm I\!R}^{m*n}$, $z_{l+1} \in {\rm I\!R}^{m*p}$ and  $z_{l} \in {\rm I\!R}^{n*p}$. The LSMR is called $p$ times with a matrix of size $n*n$ and a vector of size $n$ as inputs. So, the time complexity of this function is $O(n^3 * p )$.

\item \textbf{Complexity of \textit{Output Update} Procedure}\vspace{5pt}\\
There are two matrices among the inputs of this function which their dimensions affect the total complexity :  $W_{l} \in {\rm I\!R}^{m*n}$ and  $x_{l-1} \in {\rm I\!R}^{n*p}$. The most expensive computation in this procedure is the multiplication of these two matrices. Therefore the computational complexity of this function is $O(m*n*p)$.
\item \textbf{Complexity of \textit{Last Output Update} Procedure:}\vspace{5pt}\\
The last output update function contains a multiplication of two matrices  $W_{L} \in {\rm I\!R}^{OS*HS}$ and  $x_{L-1} \in {\rm I\!R}^{HS*N}$ with computational complexity of $O(OS* HS* N )$ .
\item \textbf{Complexity of \textit{Lagrangian Update} Procedure:}\vspace{5pt}\\
The complexity of this function is computed the same as the last output update function.
\item \textbf{Complexity Analysis of Training Neural Networks:} \vspace{5pt}\\
In each iteration of training a neural network with ADMM, weight update, activation update and output update are called for every layer. Here, we assume that all hidden layers have identical number of nodes and this number of nodes are less than the number of training samples. We calculated the complexity of major parts against hidden size $HS$, number $N$ and dimensionality $D$ of input data and dimensionality of output $OS$. The results for two-layer, three-layer and four-layer networks can be found in tables \ref{tab:twolayer}, \ref{tab:threelayer} and \ref{tab:fourlayer} respectively.
\end{itemize}
\begin{table}
\def\arraystretch{1.5}%
\caption{Complexity analysis of two layer network}
\begin{center}
\label{tab:twolayer}
\begin{tabular}[h]{ |l | l| c| } 
\hline
Layer & Procedure & Complexity \\
  \hline
  \multirow{3}{8em}{1 (Input Layer)} & Weight Update & $HS*N*D^2$  \\ 
    \cline{2-3} 
  & Activation Update & $HS^3*N$ \\ 
    \cline{2-3}
  & Output Update & $HS*N*D$ \\ 
  \hline
    \multirow{3}{8em}{2 (Output Layer)} & Weight Update & $HS^2*N*OS$  \\ 
    \cline{2-3} 
  & Last Output Update & $HS*N*OS$ \\ 
    \cline{2-3}
  & Lagrangian Update & $HS*N*OS$\\ 
  \hline

\end{tabular}
\end{center}
\end{table}
\begin{table}
\def\arraystretch{1.5}%
\caption{Complexity analysis of three layer network}
\begin{center}
\label{tab:threelayer}
\begin{tabular}[h]{ |l | l| c| } 
\hline
Layer & Procedure & Complexity \\
  \hline
  \multirow{3}{8em}{1 (Input Layer)} & Weight Update &  $HS*N*D^2$  \\ 
    \cline{2-3} 
  & Activation Update &  $HS^3*N$ \\ 
    \cline{2-3}
  & Output Update &  $HS*N*D$  \\ 
  \hline
    \multirow{3}{8em}{2 (Hidden Layer)} & Weight Update & $HS^3*N$  \\ 
    \cline{2-3} 
  & Activation Update & $HS^3*N$ \\ 
    \cline{2-3}
  & Output Update & $HS^2*N$ \\ 
  \hline
    \multirow{3}{8em}{3 (Output Layer)} & Weight Update & $HS^2*N*OS$ \\ 
    \cline{2-3} 
  &Last Output Update & $HS*N*OS$ \\ 
    \cline{2-3}
  & Lagrangian Update & $HS*N*OS$ \\ 
  \hline
  
\end{tabular}
\end{center}
\end{table}

\begin{table}
\def\arraystretch{1.5}%
\caption{Complexity analysis of four layer network}
\begin{center}
\label{tab:fourlayer}
\begin{tabular}[h]{ |l | l| c| } 
\hline
Layer & Procedure & Complexity \\
  \hline
  \multirow{3}{8em}{1 (Input Layer)} & Weight Update & $HS*N*D^2$  \\ 
    \cline{2-3} 
  & Activation Update &$HS^3*N$\\ 
    \cline{2-3}
  & Output Update & $HS*N*D$\\ 
  \hline
    \multirow{3}{8em}{2 (Hidden Layer)} & Weight Update &$HS^3*N$  \\ 
    \cline{2-3} 
  & Activation Update & $HS^3*N$\\ 
    \cline{2-3}
  & Output Update & $HS^2*N$ \\ 
  \hline
    \multirow{3}{8em}{3 (Hidden Layer)} & Weight Update & $HS^3*N$  \\ 
    \cline{2-3} 
  & Activation Update & $HS^3*N$ \\ 
    \cline{2-3}
  & Output Update &$HS^2*N$ \\ 
  \hline
      \multirow{3}{8em}{4 (Output Layer)} & Weight Update & $HS^2*N*OS$ \\ 
    \cline{2-3} 
  &Last Output Update & $HS*N*OS$ \\ 
    \cline{2-3}
  & Lagrangian Update & $HS*N*OS$ \\ 
  \hline
\end{tabular}
\end{center}
\end{table}

\chapter{Experimental Results}
\label{ch4}
In this project, two datasets were used for experiments. IRIS  \cite{fisher1950use} and subset of a more difficult dataset, HIGGS \cite{higgs}. The experiments include comparing the test accuracy of a neural network using our implemented method against two state-of-the-art gradient-based methods and also measuring the execution time of different procedures in the suggested method. We can summarise the key observations as the following: 
\begin{itemize}
    \item Avoiding matrix inversion by use of LSMR does not have a significant impact on the test accuracy of ADMM algorithm.
    \item We have achieved better accuracy compared to both SGD and Adam on HIGGS and IRIS datasets with small-sized feed-forward neural networks.
\end{itemize}
All the experiments were done on the following platform and software:\\
OS: macOS Catalina version 10.15.2\\
Processor: 2.7 GHz Dual-Core Intel Core i5\\
Memory: 8 GB 1867 MHz DDR3\\
Python version 3.7.4 \\
Pytorch version 1.4.0

\section{Experiment Setup}
In our experiments, we aimed to compare the test accuracy of the implemented method versus two gradient-based methods. The first method is Stochastic Gradient Descent which is one of the most primary methods and therefore, it is a simple and acceptable baseline. The other compared method is Adam which is at the moment one of the most popular gradient-based methods due to its computational efficiency and ease of tuning \cite{kingma2014adam}.\vspace{8pt}\\
The compared neural networks that use SGD and Adam as their optimiser have been implemented using Pytorch library \cite{NEURIPS2019_9015} with all the hyperparameters being default values.\vspace{8pt}\\
As described in section \ref{ADMMNNsection}, there are two penalty parameters to be set in our ADMM-based implementation. In all experiments reported here $\gamma_i = 10$ and $\beta_i = 1$. Also $x_l$, $z_l$ and $W_l$ matrices were initialized using i.i.d Gaussian random variables.

\section{Results}
Firstly, we compared ADMM and ADMM-LSMR in order to inspect the effect of using iterative least-square methods on the test accuracy. Secondly, we compared ADMM-LSMR against the gradient-based methods. 
\subsection{Experiments on HIGGS}
We have used this dataset both to compare ADMM versus ADMM-LSMR, and to compare ADMM-LSMR versus SGD and Adam. Each algorithm has been run 200 times in this experiment set and the architecture of neural networks used was a four-layer network with hidden size of 28.
In our experiment, all distributions of test accuracies on this dataset turned out to be normal. So in order to compare them we were able to use t-test.
\subsubsection{ADMM versus ADMM-LSMR}
\begin{figure}[H]
\centering
\includegraphics[width = 0.7\hsize]{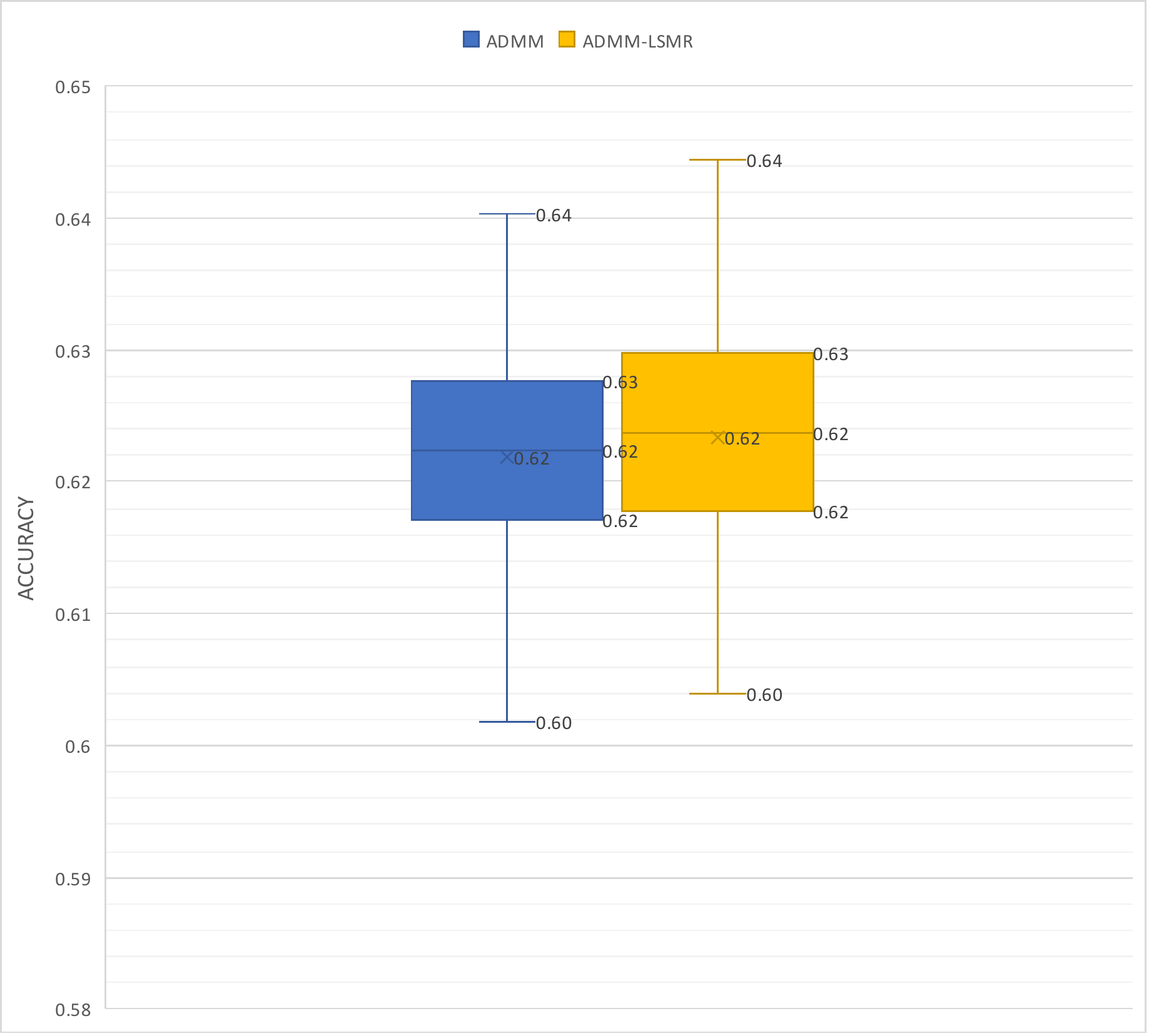}
\caption{Box plot of test accuracies on HIGGS dataset. ADMM versus ADMM-LSMR}
\label{fig:ADMM-HIGGS}
\end{figure}

\begin{table}[H]
\def\arraystretch{1.5}%
\caption{Comparing ADMM versus ADMM-LSMR on HIGGS dataset}
\begin{center}
\label{tab:ADMM-HIGGS}
\begin{tabular}{l|l|l|}
\cline{2-3}
                                 & Mean & STDV \\ \hline
\multicolumn{1}{|l|}{ADMM-LSMR}  &  0.6219    &  0.0080    \\ \hline
\multicolumn{1}{|l|}{ADMM}       &  0.6234    &  0.0081    \\ \hline
\end{tabular}
\end{center}
\end{table}

\definecolor{backcolour}{rgb}{0.95,0.95,0.95}
\lstdefinestyle{mystyle}{
  backgroundcolor=\color{backcolour},   commentstyle=\color{green},
  numberstyle=\tiny\color{gray},
  stringstyle=\color{purple},
  basicstyle=\ttfamily\footnotesize,
  breakatwhitespace=false,         
  breaklines=true,                 
  captionpos=b,                    
  keepspaces=true,                 
  numbers=left,                    
  numbersep=5pt,                  
  showspaces=false,                
  showstringspaces=false,
  showtabs=false,                  
  tabsize=2
}

\lstset{style=mystyle}
\begin{lstlisting}[language=R, caption= T-test for comparing ADMM versus ADMM-LSMR on HIGGS dataset ,label=tadmm]
	Welch Two Sample t-test

data:  ADMMLSMR and ADMM
t = -1.8033, df = 397.96, p-value = 0.07209
alternative hypothesis: true difference in means is not equal to 0
99 percent confidence interval:
 -0.0035599066  0.0006362702
sample estimates:
mean of x mean of y 
0.6219455 0.6234073 

\end{lstlisting}
 \vspace{8pt}

As it is evident from figure \ref{fig:ADMM-HIGGS}, table \ref{tab:ADMM-HIGGS} and t-test result \ref{tadmm}, we can not observe a significant difference between these two distributions and the mean accuracy of ADMM-LSMR is just 0.2\% less than ADMM. We can conclude that we are able to avoid matrix inversion without sacrificing test accuracy. This is particularly important because the avoidance of matrix inversion is a big step towards hardware implementation of this method for training neural networks.
\newpage
\subsubsection{ADMM-LSMR versus SGD and Adam}
\begin{figure}[H]
\centering
\includegraphics[width = 0.7\hsize]{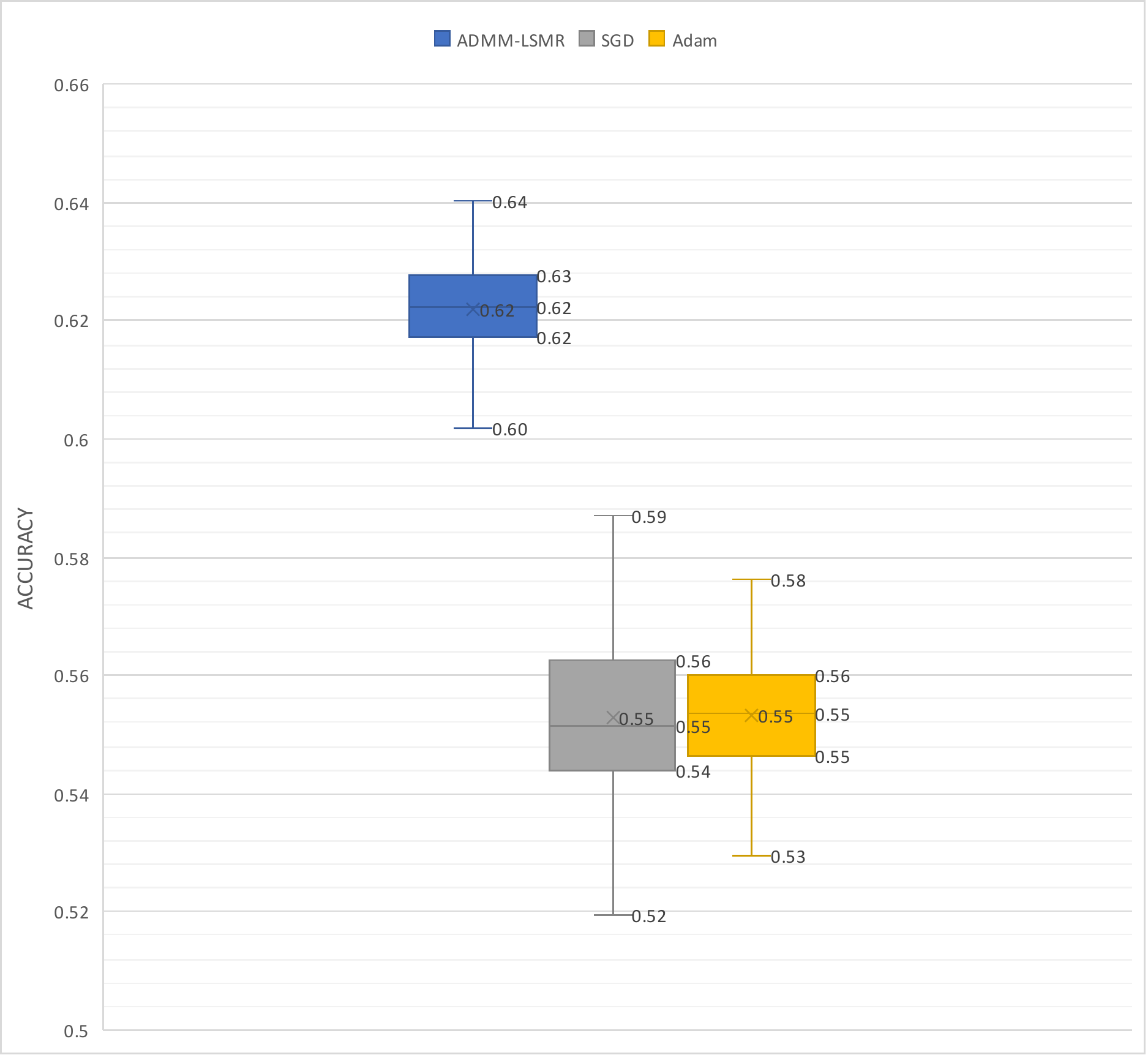}
\caption{\footnotesize Box plot of test accuracies on HIGGS dataset. ADMM-LSMR versus SGD and Adam}
\label{fig:HIGGS}
\end{figure}

\begin{table}[H]
\def\arraystretch{1.5}%
\caption{Comparing accuracies of different methods on HIGGS dataset}
\begin{center}
\label{tab:ACCURACY-HIGGS}
\begin{tabular}{l|l|l|}
\cline{2-3}
                           & Mean & STDV \\ \hline
\multicolumn{1}{|l|}{ADMM-LSMR} & 0.6219   & 0.0080  \\ 
\multicolumn{1}{|l|}{SGD}  & 0.5527   & 0.0134     \\ 
\multicolumn{1}{|l|}{Adam} & 0.5532   & 0.0105     \\ \hline
\end{tabular}
\end{center}
\end{table}
\begin{lstlisting}[language=R, caption= T-test for comparing ADMM-LSMR versus SGD on HIGGS dataset , label =tadmmsgd]
	Welch Two Sample t-test

data:  ADMMLSMR and SGD
t = 62.556, df = 326.38, p-value < 2.2e-16
alternative hypothesis: true difference in means is not equal to 0
99 percent confidence interval:
 0.06633211 0.07206425
sample estimates:
mean of x mean of y 
0.6219455 0.5527473 

\end{lstlisting}
\begin{lstlisting}[language=R, caption=T-test for comparing ADMM-LSMR versus Adam on HIGGS dataset , label =tadmmadam]
	Welch Two Sample t-test

data:  ADMMLSMR and Adam
t = 73.32, df = 373.04, p-value < 2.2e-16
alternative hypothesis: true difference in means is not equal to 0
99 percent confidence interval:
 0.06625302 0.07110334
sample estimates:
mean of x mean of y 
0.6219455 0.5532673 

\end{lstlisting}

It can be observed from figure \ref{fig:HIGGS}, table \ref{tab:ACCURACY-HIGGS} and t-test results \ref{tadmmsgd} and \ref{tadmmadam} that the accuracy of ADMM-LSMR is significantly higher than both SGD and Adam. The mean accuracy of ADMM-LSMR is  6.9\% and 6.8\% better than SGD and Adam respectively, which is a very promising achievement.
\\

\subsection{Experiments on IRIS}
We have used this dataset to compare ADMM-LSMR versus SGD and Adam with 1000 runs for each algorithm.  The architecture of neural networks used in this set of experiments was a three-layer network with hidden size of 8.
\subsubsection{ADMM-LSMR versus SGD and Adam}
\begin{figure}[H]
\centering
\includegraphics[width = 0.7\hsize]{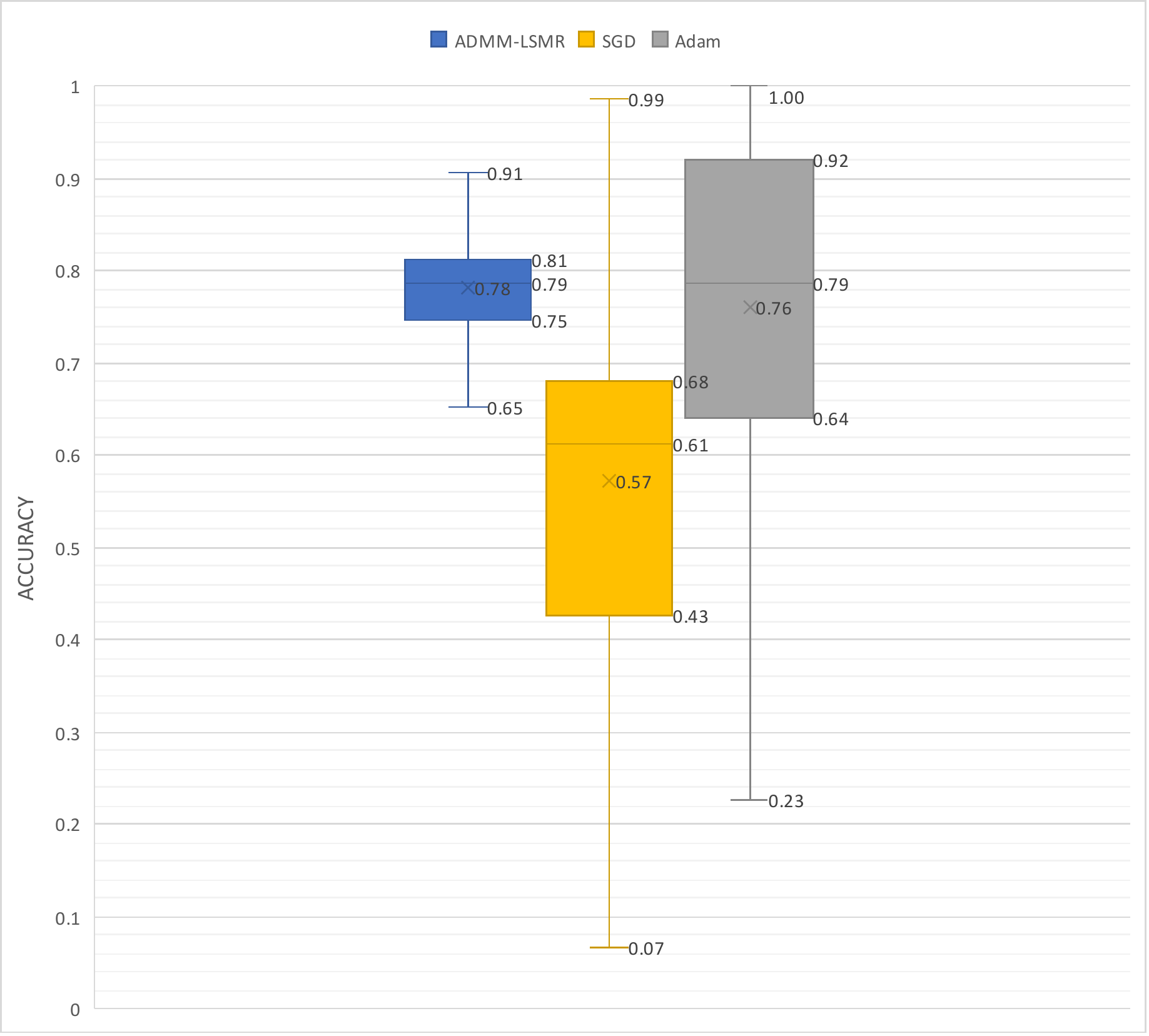}
\caption{ \footnotesize Box plot of test accuracies on IRIS dataset. ADMM-LSMR versus SGD and Adam}
\label{fig:IRIS}
\end{figure}

\begin{table}[H]
\def\arraystretch{1.5}%
\caption{Comparing accuracies of different methods on IRIS dataset. }
\begin{center}
\label{tab:ACCURACY-IRIS}
\begin{tabular}{l|l|l|}
\cline{2-3}
                                 & Mean & STDV \\ \hline
\multicolumn{1}{|l|}{ADMM-LSMR}  & 0.7826  &   0.0556   \\ 
\multicolumn{1}{|l|}{SGD}        & 0.5722  &   0.1873  \\ 
\multicolumn{1}{|l|}{Adam}       & 0.7599  &   0.1838   \\ \hline
\end{tabular}
\end{center}
\end{table}
Based on figure \ref{fig:IRIS} and table \ref{tab:ACCURACY-IRIS} we can observe that the mean accuracy of ADMM-LSMR is higher than both Adam and SGD. The mean accuracy of ADMM-LSMR is  21.0\% and 2.2\% more than SGD and Adam respectively. While the difference between ADMM-LSMR and Adam is small, we can infer that ADMM-LSMR is more consistent having much lower standard deviation. 

\subsection{Run Time Measurements}
In this section, we provide the results of measuring the execution time for each procedure of the implemented method. For this purpose, we used a three-layer neural network on IRIS dataset and we increased the hidden size from 5 to 100. Each point is the result of averaging the execution time for 10 different runs. In the following figures, execution time versus hidden size could be found for each entry of the table \ref{tab:threelayer}.
\subsubsection{Input Layer - Weight Update }

\begin{figure}[H]
\centering
\includegraphics[width = 0.5\hsize]{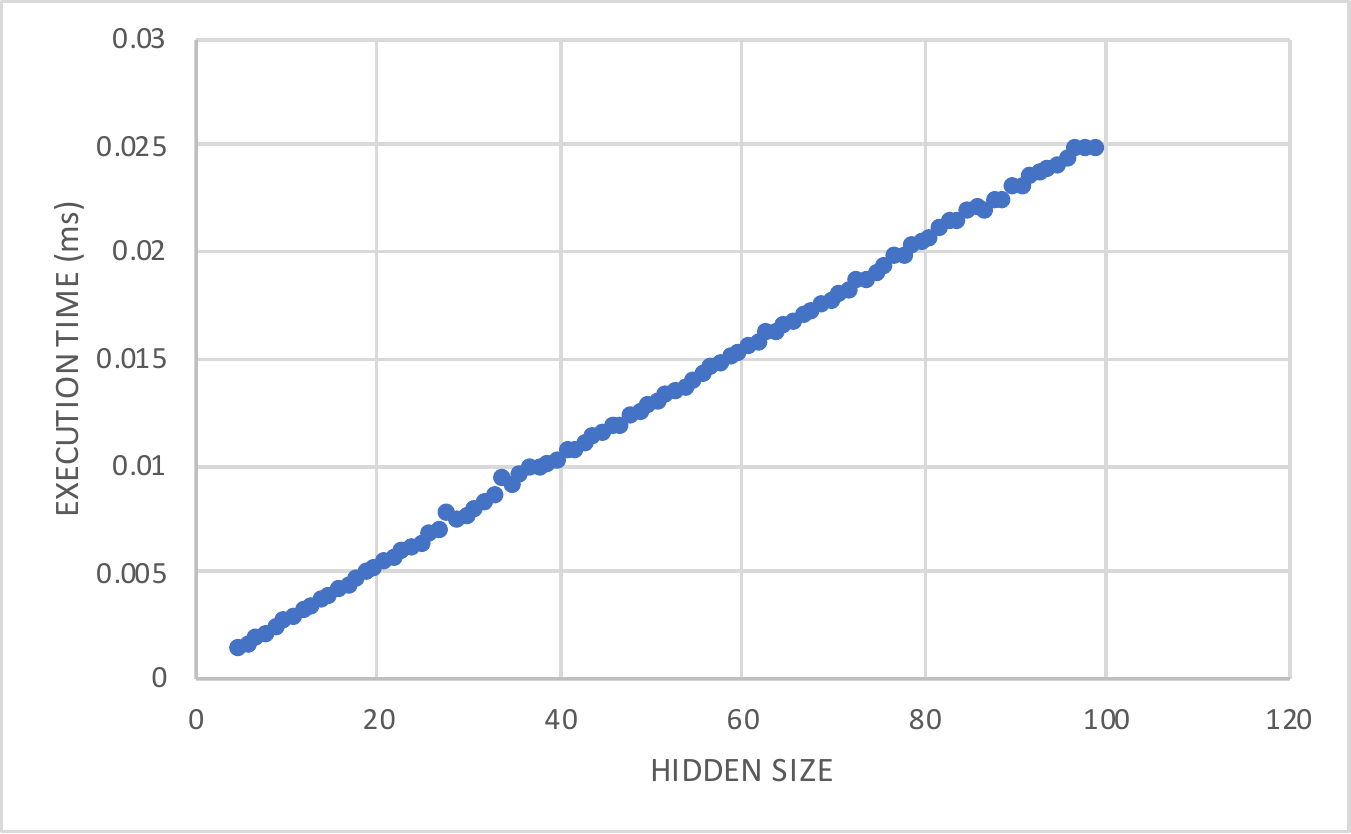}
\caption{ \footnotesize Execution time of weight update procedure of input layer }
\end{figure}

\subsubsection{Input Layer - Activation Update }
\begin{figure}[H]
\centering
\includegraphics[width = 0.5\hsize]{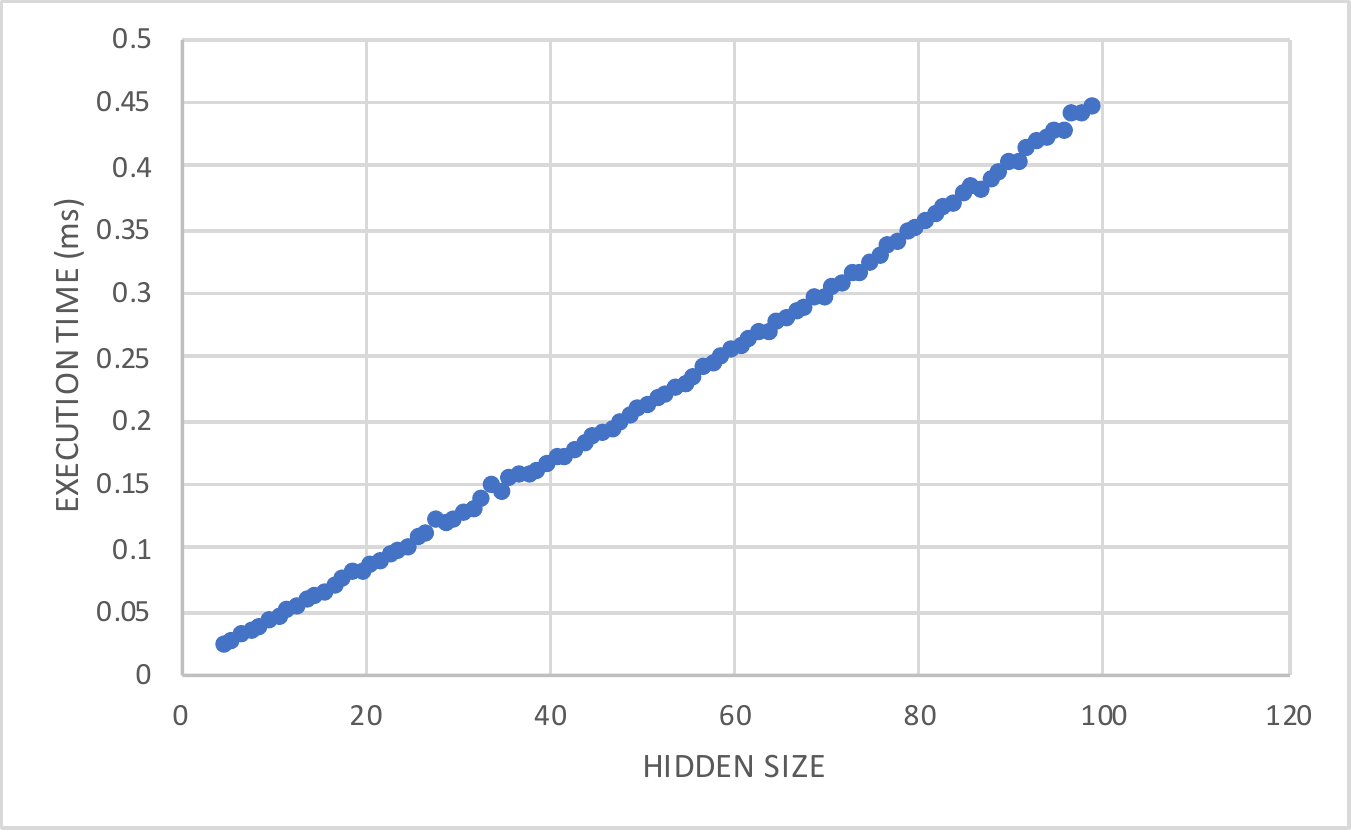}
\caption{ \footnotesize Execution time of activation update procedure of input layer }
\end{figure}

\subsubsection{Input Layer - Output Update }
\begin{figure}[H]
\centering
\includegraphics[width = 0.5\hsize]{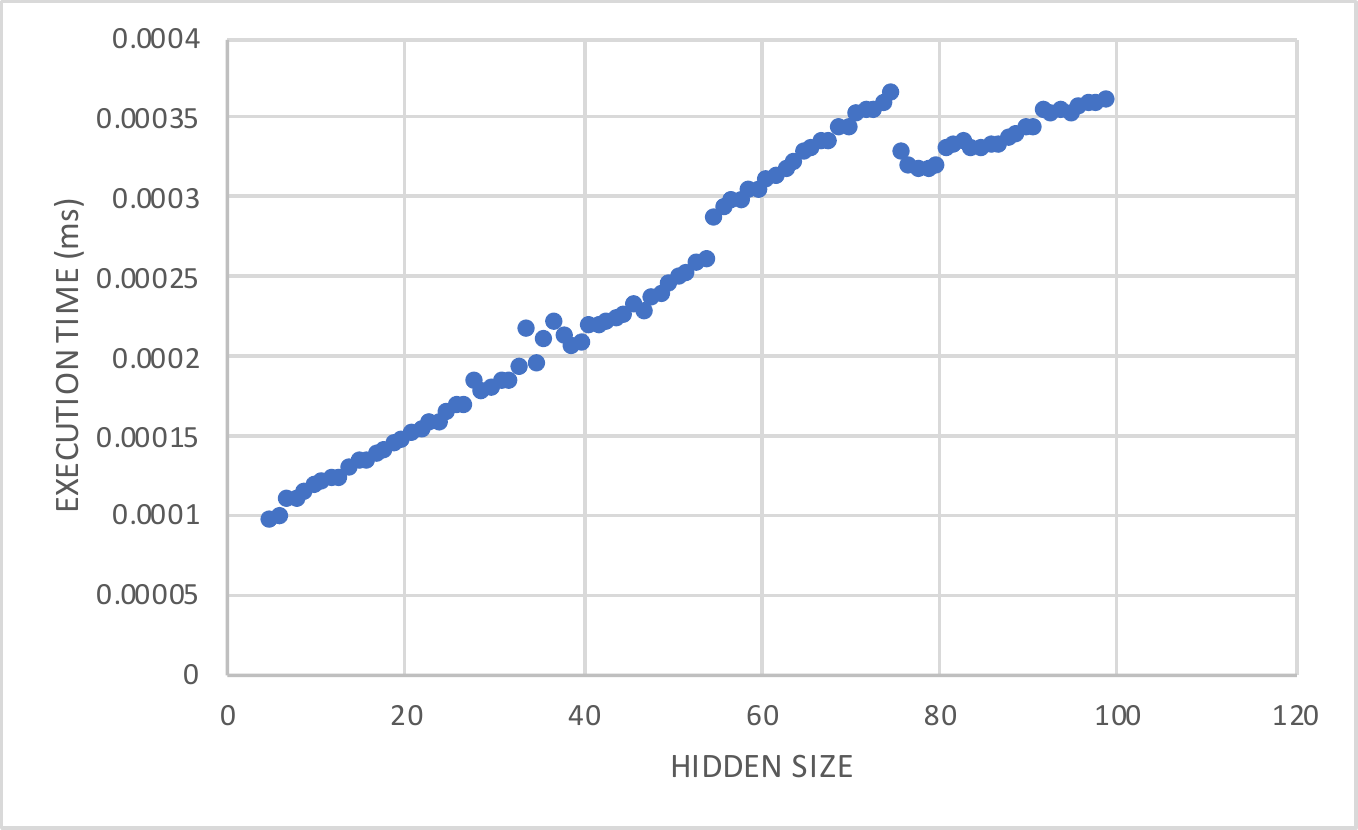}
\caption{ \footnotesize Execution time of output update procedure of input layer }
\end{figure}

\subsubsection{Hidden Layer - Weight Update }
\begin{figure}[H]
\centering
\includegraphics[width = 0.5\hsize]{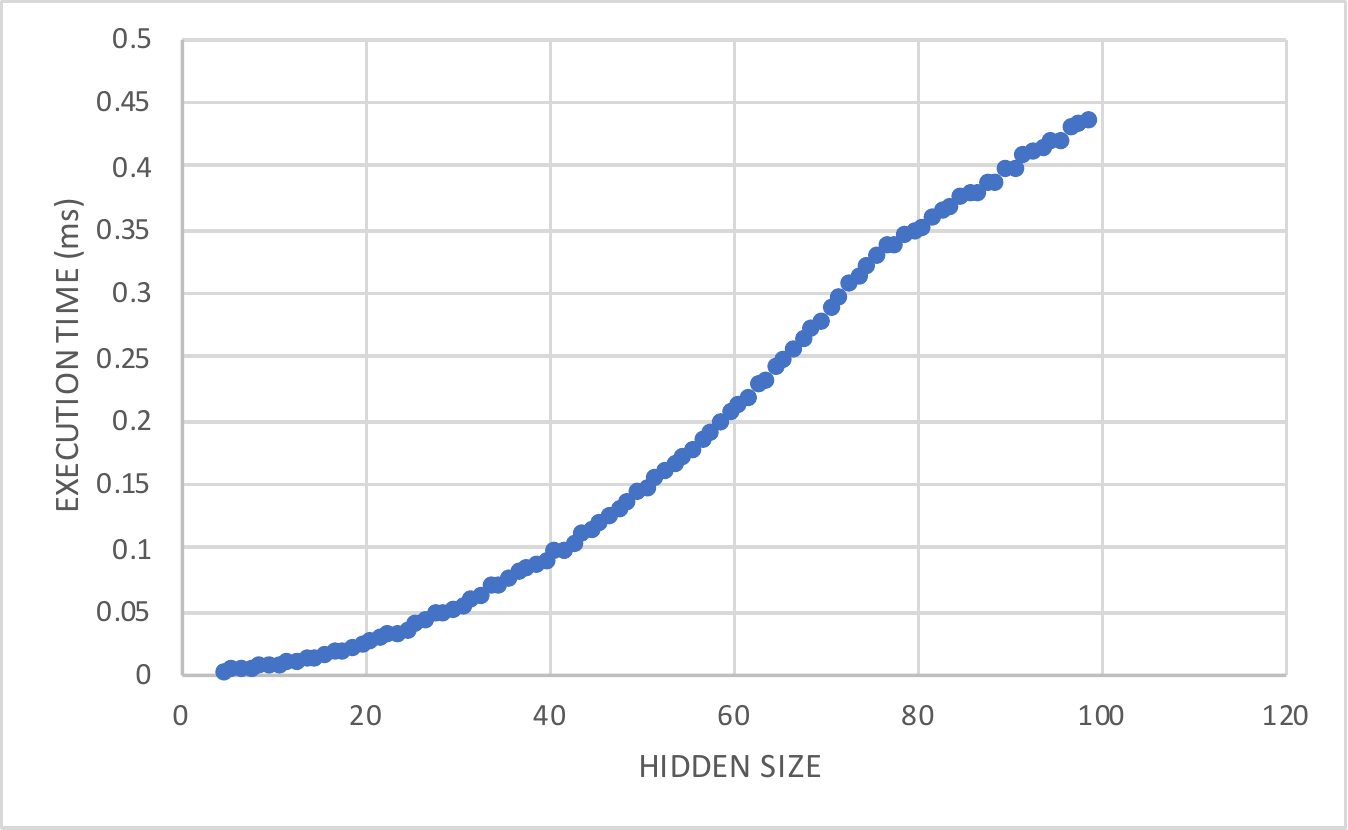}
\caption{ \footnotesize Execution time of weight update procedure of hidden layer }
\end{figure}

\subsubsection{Hidden Layer - Activation Update }
\begin{figure}[H]
\centering
\includegraphics[width = 0.5\hsize]{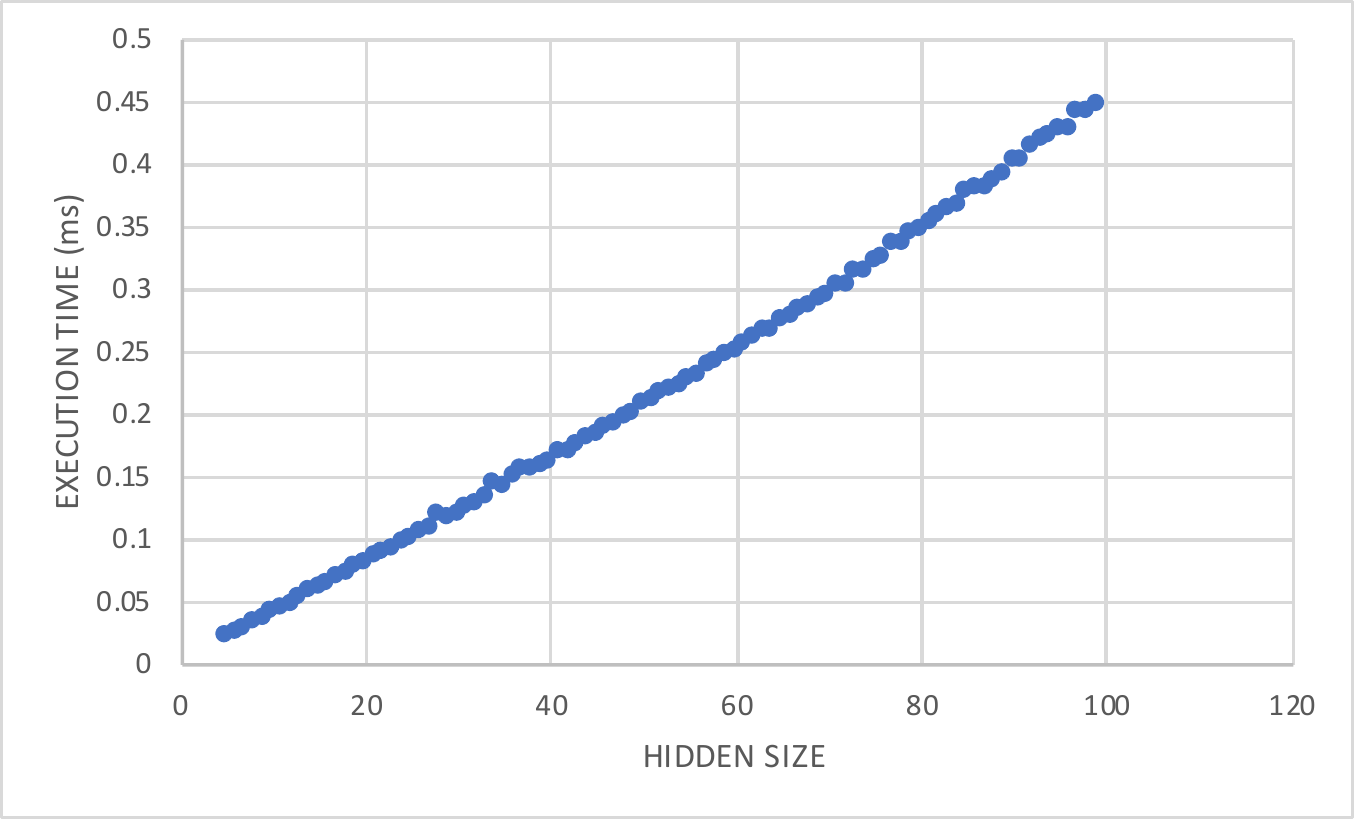}
\caption{ \footnotesize Execution time of activation update procedure of hidden layer }
\end{figure}

\subsubsection{Hidden Layer - Output Update }
\begin{figure}[H]
\centering
\includegraphics[width = 0.5\hsize]{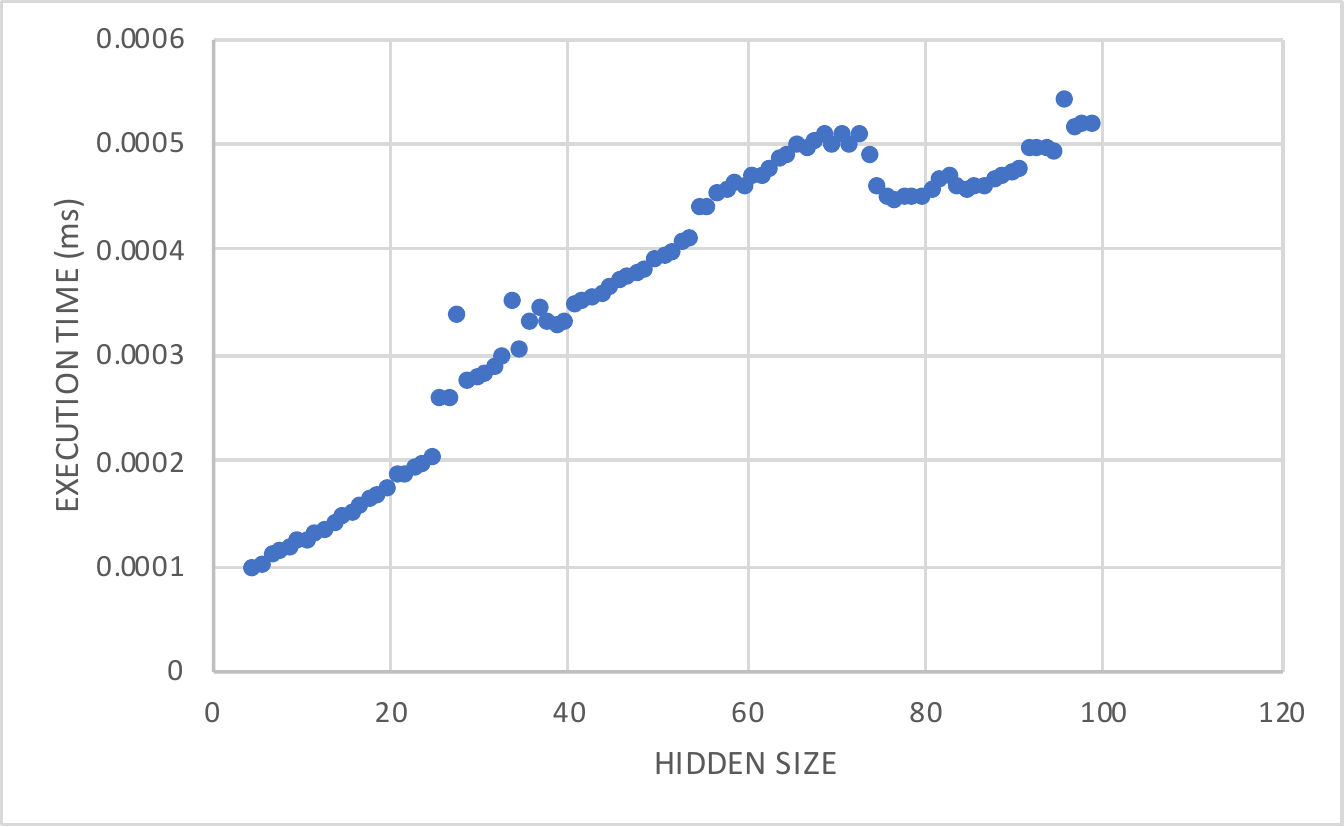}
\caption{ \footnotesize Execution time of output update procedure of hidden layer }
\end{figure}

\subsubsection{Output Layer - Weight Update }
\begin{figure}[H]
\centering
\includegraphics[width = 0.5\hsize]{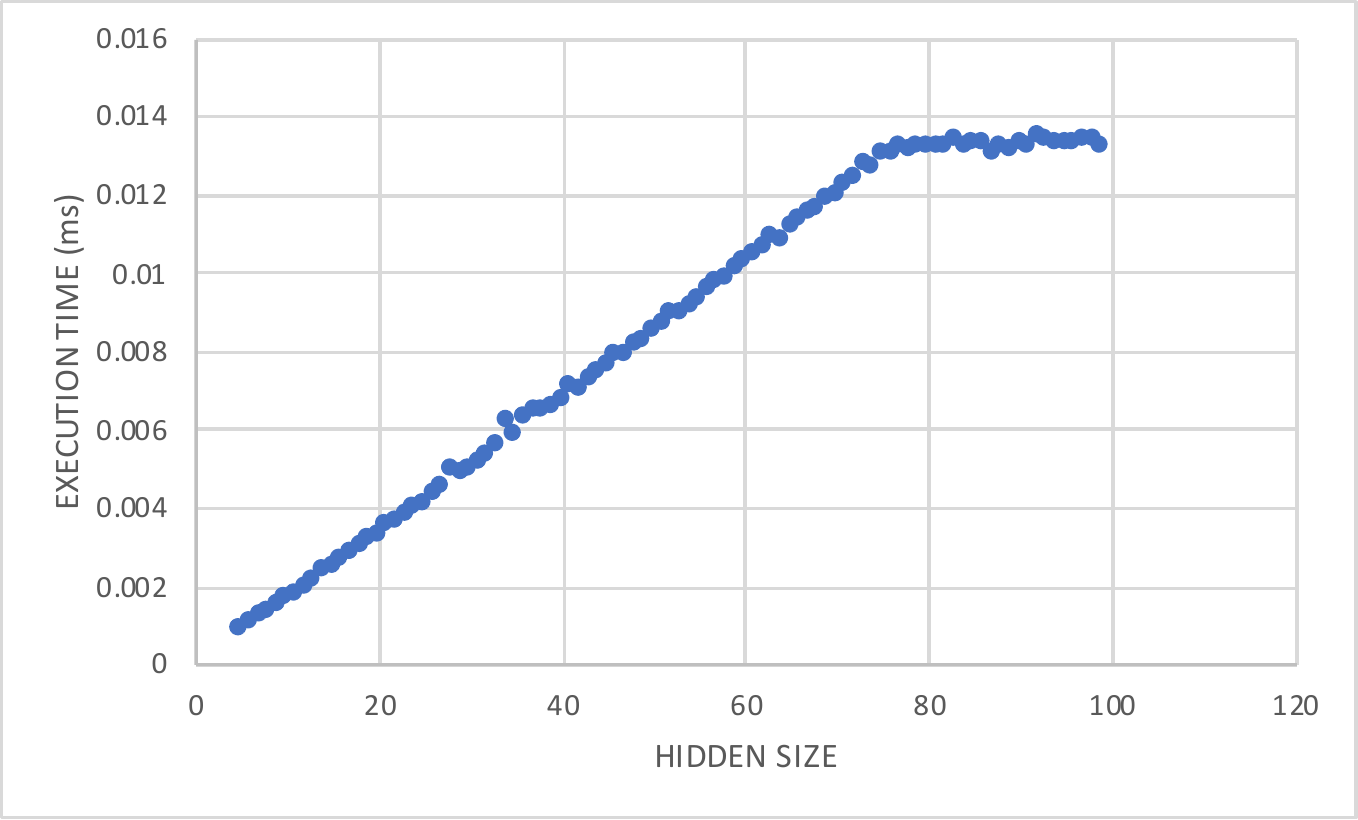}
\caption{ \footnotesize Execution time of weight update procedure of output layer }
\end{figure}
\subsubsection{Output Layer - Last Output Update }
\begin{figure}[H]
\centering
\includegraphics[width = 0.5\hsize]{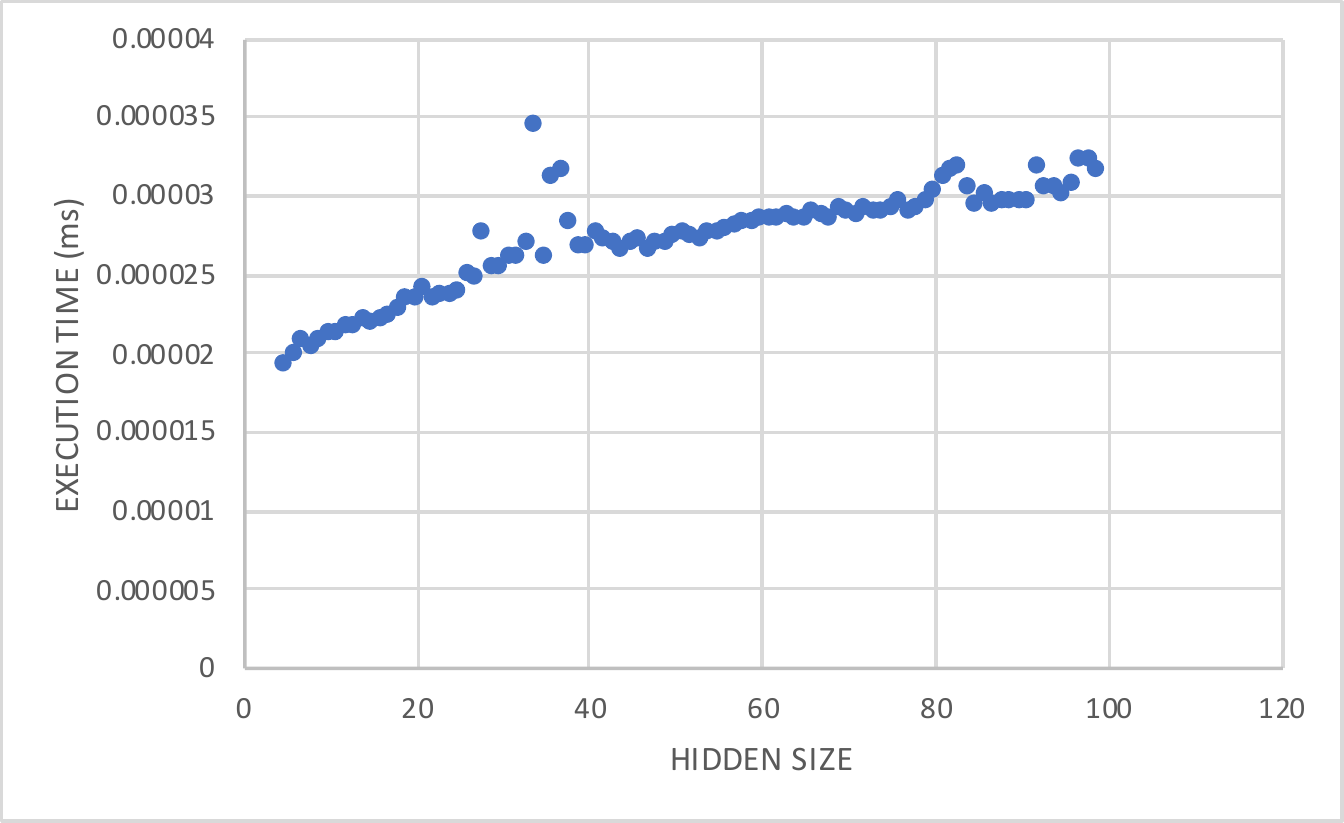}
\caption{ \footnotesize Execution time of last output update update procedure of output layer }
\end{figure}

\subsubsection{Output Layer - Lagrangian Update }
\begin{figure}[H]
\centering
\includegraphics[width = 0.5\hsize]{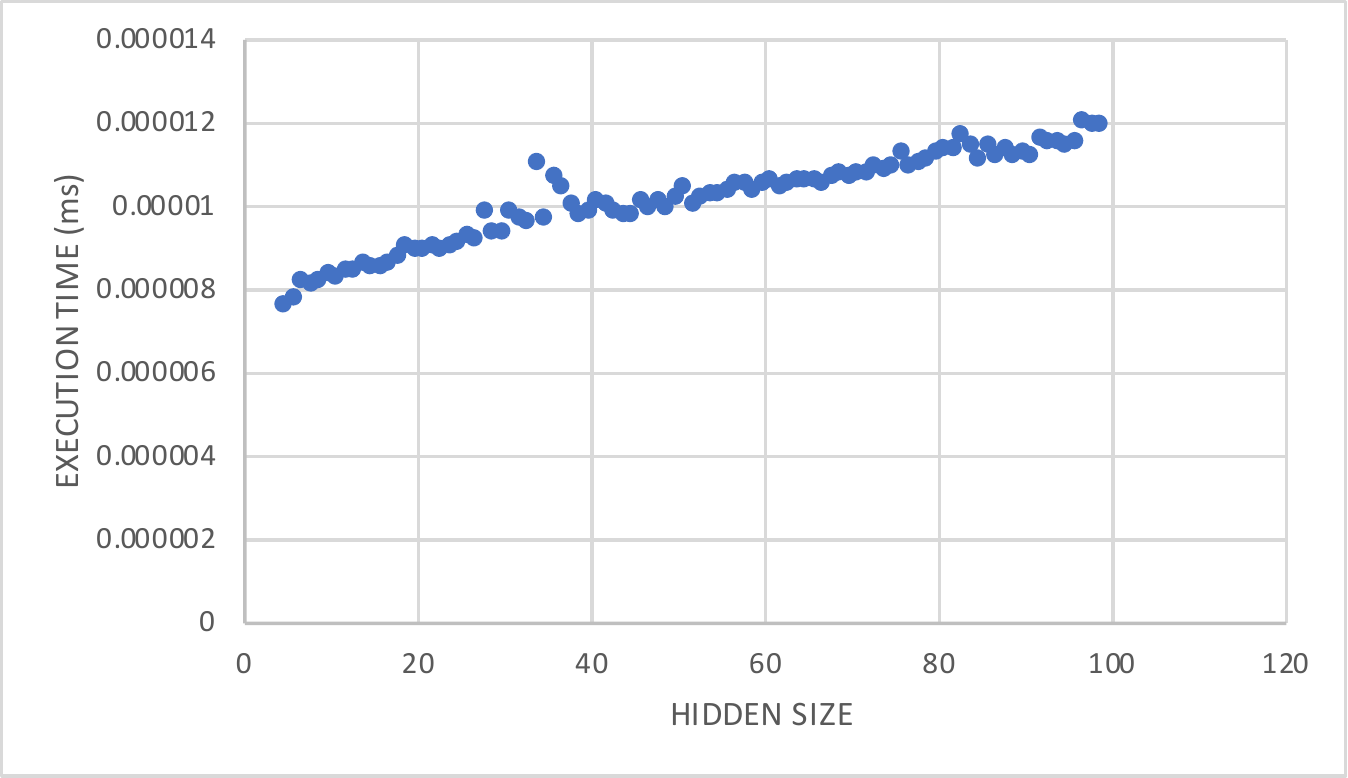}
\caption{ \footnotesize Execution time of Lagrangian update procedure of output layer }
\end{figure}

As it is evident from the plots, the behaviour of the execution time of most of the procedures differ from and are better than the analysis provided in section \ref{complexity}. One possible explanation could be that Numpy and Python perform heavy optimisations that could affect the execution times. The other possibility is that for the procedures which their execution time is very short, the numbers are more affected with the noise of measurement.
\chapter{Conclusion and Future Work}
\label{ch5}

In this project, we implemented an algorithm for training feed-forward neural networks based on ADMM. 
We altered the suggested implementation in \cite{taylor2016training} and used iterative least-square methods as a replacement for computing matrix inversion. It is observed that ADMM-based neural networks are significantly more performant than gradient-based methods. Also as it is evident from the results the use of LSMR, which is an iterative least-square method, does not have a significant effect on the accuracy of the ADMM-based neural networks while making it suitable for hardware acceleration.  \vspace{8pt}\\
There are multiple characteristics which make our implemented method hardware compatible. First of all, as it is not a gradient-based method, it does not suffer from sequential dependency and can be parallelised more efficiently. Second, it is more suitable than gradient-based methods to be used alongside low-precision and fixed-point numbers which is a common approach in hardware implementation of training methods. While sacrificing precision is a known issue associated with using low-precision numbers in gradient-based methods, our implemented algorithm, as an ADMM-based method, can evade such a loss by avoiding back-propagation.\vspace{8pt}\\
Our experiments were limited to small-sized feed-forward neural networks. As further work, the performance of ADMM in larger neural networks and also in recurrent and convolutional neural networks can be explored. Furthermore, the utilization of other activation functions can be investigated. Ultimately, implementing a hardware-accelerated version of the algorithm (possibly on FPGAs) and exploiting the characteristics of the method for parallelism could lead to very promising results.


\bibliography{main}
\bibliographystyle{ieeetr}
\end{document}